\documentclass[a4paper, 10pt]{paper}      
                                                          
\usepackage{amsmath}
\usepackage{amssymb}
\usepackage[usenames,dvipsnames]{color}
\usepackage[ruled,vlined]{algorithm2e}
\usepackage{algorithmic}
\usepackage{graphicx}
\usepackage{subfig}
\usepackage[utf8]{inputenc}
\usepackage[T1]{fontenc}

\providecommand{\abs}[1]{\lvert #1 \rvert}

\renewcommand{\Re}{{\mathbb{R}}}
\def\ee{{\boldsymbol{e}}}
\def\zero{{\boldsymbol{0}}}
\def\one{{\boldsymbol{1}}}
\def\xx{{\boldsymbol{x}}}

\def\ww{{\boldsymbol{w}}}
\def\aa{{\boldsymbol{a}}}
\def\acronym{{SOLO}}

\title{\LARGE \bf  Sparse Online Low-Rank Projection and Outlier Rejection (SOLO) \\
for 3-D Rigid-Body Motion Registration}

\author{Chris Slaughter, Allen Y. Yang, Justin Bagwell, Costa Checkles,\\ Luis Sentis, and Sriram Vishwanath%
\thanks{C. Slaughter, J. Bagwell, C. Checkles and S. Vishwanath are with Electrical and Computer Engineering Department, University of Texas, Austin, USA.
	{\tt\small <chris.c.slaughter@gmail.com, justindbagwell@mail.utexas.edu, ccheckles@utexas.edu, sriram@austin.utexas.edu>}.
	A. Yang is with the EECS Department, University of California, Berkeley, USA.
        {\tt\small <yang@eecs.berkeley.edu>}.
        L. Sentis is with the Mechanical Engineering Depatment, University of Texas, Austin, USA.
        {\tt\small <lsentis@austin.utexas.edu>}.
        This work was supported in part by the ONR, an Intel Graduate Fellowship, NSF CNS-0905200, ARO MURI W911NF-06-1-0076, and a Willow Garage gift.}        
}

\begin{document}

\maketitle
\thispagestyle{empty}
\pagestyle{empty}

\begin{abstract}
Motivated by an emerging theory of robust low-rank matrix representation, in this paper, we introduce a novel solution for online rigid-body motion registration. The goal is to develop algorithmic techniques that enable a robust, real-time motion registration solution suitable for low-cost, portable 3-D camera devices. Assuming 3-D image features are tracked via a standard tracker, the algorithm first utilizes Robust PCA to initialize a low-rank shape representation of the rigid body. Robust PCA finds the global optimal solution of the initialization, while its complexity is comparable to singular value decomposition. In the online update stage, we propose a more efficient algorithm for sparse subspace projection to sequentially project new feature observations onto the shape subspace. The lightweight update stage guarantees the real-time performance of the solution while maintaining good registration even when the image sequence is contaminated by noise, gross data corruption, outlying features, and missing data. The state-of-the-art accuracy of the solution is validated through extensive simulation and a real-world experiment, while the system enjoys one to two orders of magnitude speed-up compared to well-established RANSAC solutions. The new algorithm will be released online to aid peer evaluation.
\end{abstract}

\section{Introduction}

\emph{Rigid body motion registration} (RBMR) is one of the fundamental problems in machine vision and robotics. Given a dynamic scene that contains a (dominant) rigid body object and a cluttered background, certain salient image feature points can be extracted and tracked with considerable accuracy across multiple image frames \cite{ShiJ1994-CVPR}. The task of RBMR then involves identifying the image features that are associated only with the rigid-body object in the foreground and subsequently recovering its rigid-body transformation across multiple frames. Traditionally, RBMR has been mainly conducted in 2-D image space, with the assumption of the camera projection model from simple orthographic projection \cite{TomasiC1992-IJCV} to more realistic camera models such as paraperspective \cite{PoelmanC1997-PAMI} and affine \cite{KahlF1999-IJCV}. In problems such as RBMR, Structure from Motion (SfM), and motion segmentation \cite{KanataniK2001-ICCV,VidalR2006}, a fundamental observation is that a data matrix that contains the coordinates of tracked image features in column form can be factorized as a camera matrix that represents the motion and a shape matrix that represents the shape of the rigid body in the world coordinates. Furthermore, if the data are noise-free, then the feature vectors in the data matrix lie in a 4-D subspace, as the rank of the shape matrix in the world coordinates is at most four \cite{TomasiC1992-IJCV}.

In practice, the RBMR problem can become more challenging if the tracked image features are perturbed by moderate noise, gross image corruption (e.g., when the features are occluded), and missing data (e.g., when the features leave the field of view). In robust statistics, it is well known that the optimal solution to recover a subspace model when the data is complete yet affected by Gaussian noise is \emph{singular value decomposition} (SVD). Solving other image nuisances caused by gross measurement error corresponds to the problem of robust estimation of a low-dimensional subspace model in the presence of corruption and missing data. In \cite{HartleyR2003}, for instance, the issue of missing data was addressed by robustifying SVD via Power Factorization. In \cite{ChenP2004-PAMI}, the same issue was addressed by an iterative imputation strategy. 

In the case of outlier rejection, arguably the most popular robust model estimation algorithm in computer vision is Random Sample Consensus (RANSAC) \cite{FischlerM1981}. In the context of RBMR, the standard procedure of RANSAC is  to apply the iterative \emph{hypothesize-and-verify} scheme on a frame-by-frame basis to recover rigid-body motion \cite{TorrP2003-PAMI,StewartC1999-SIAM,WangH2004-IJCV}. In the context of dimensionality reduction, RANSAC can also be applied to recover low-dimensional subspace models \cite{YangA2006-CVPR}, such as the above shape model in motion registration. 

Nevertheless, the aforementioned solutions have two major drawbacks. In the case of missing data, methods such as Power Factorization or incremental SVD cannot guarantee the global convergence of the estimate \cite{HartleyR2003,ChenP2004-PAMI}. In the case of outlier rejection, the RANSAC procedure is known to be  expensive to deploy in a real-time, online fashion, such as in the solutions for \emph{simultaneous localization and mapping} (SLAM) \cite{WilliamsB2007-ICRA,SaekiK2009-ICRA}. Therefore, a better solution than the state of the art should provide provable global optimality to compensate missing data, image corruption, and erroneous feature tracks, and at the same time should be more efficient to recover rigid body motion from a video sequence in a online fashion. In this paper, we propose a highly robust solution to address this problem. 

\subsection{Contributions}
Our solution is motivated by the emerging theory of Robust PCA (RPCA) \cite{CandesE2009-ACM,ZhouZ2010}. In particular, RPCA provides a unified solution to estimating low-rank matrices in the cases of both missing data \emph{and} random data corruption \cite{CandesE2009-ACM}. The algorithm is guaranteed to converge to the global optimum if the ambient space dimension is sufficiently high. Compared to other existing solutions such as incremental SVD and RANSAC, the set of heuristic parameters one needs to tune is also minimal. Furthermore, recent progress in convex optimization has led to very efficient numerical implementation of RPCA with the computational complexity comparable to that of classical SVD \cite{LinZ2009}. 

Our proposed solution to online 3-D motion registration consists of two steps. In the initialization step, RPCA is used to estimate a low-rank representation of the rigid-body motion within the first several image frames, which establishes a global shape model of the rigid body. In the online update step, we propose a sparse subspace projection method that projects new observations onto the low-dimensional shape model, simultaneously correcting possible sparse data corruption. The overall algorithm is called \emph{Sparse Online Low-rank projection and Outlier rejection} (\acronym).

Compared to the popular method of RANSAC, one major benefit of the new solution is that by enforcing a low-rank shape model, those sparsely corrupted image features can be compensated instead of simply being discarded. In this paper, we apply the algorithm to 3-D motion features that are tracked by the relatively new Microsoft Kinect motion sensor. However, the same algorithm can help address the more traditional RBMR problems with 2-D image features. Through extensive simulation and a real-world experiment, we demonstrate that SOLO solves the online RBMR problem with state-of-the-art accuracy and more importantly improved speed of one or two orders of magnitude faster than RANSAC. To aid peer evaluation, the MATLAB/C source code of our algorithm will be released on our website.

\section{3-D Feature Tracking}\label{sec:track}

\begin{figure}[th!]
  \centering
  \includegraphics[width=0.8\textwidth]{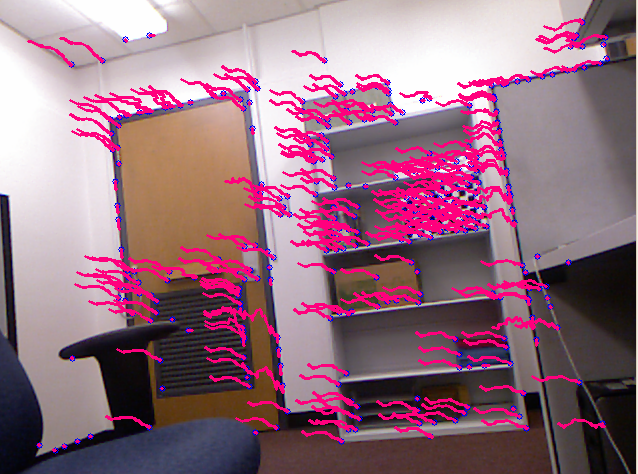}
  \caption{\small Tracking results of an indoor scene shown on the first frame of the sequence.}
  \label{fig:tracker}
\end{figure}

In this section, we briefly describe the 3-D feature tracking methodology used in this paper. In our 3-D tracking subsystem (e.g., on Microsoft Kinect), we first identify salient image features, and then track them frame by frame in image space (as an example shown in Figure \ref{fig:tracker}). The features are then reprojected onto the camera coordinate system using depth measurements. Over time, new features are extracted on periodic intervals to maintain a dense set over the image geometry.  Each feature is tracked independently, and may be dropped once it leaves the field of view or produces spurious results (jumps) in camera space.

For tracking, we use the Kanade-Lucas-Tomasi feature tracker (KLT) \cite{ShiJ1994-CVPR}.  It is well known that the KLT tracker is extremely fast and can run in real time on a standard desktop computer.  For KLT to work effectively, the extracted features must exhibit local saliency.  To achieve this and produce a dense set of features over scenes, we use the Harris corner detector as well as a Difference of Gaussians (DoG) extractor \cite{TuytelaarsT2008}.  Only the lowest two levels of the DoG pyramid are used.  This ensures that the features exhibit high local saliency in a small window and are spatially well-localized.  

One implicit advantage of tracking features across multiple frames is that it permits the tracking data to be represented naturally as a matrix.  Each (sample-indexed) row represents observations of multiple features in a single time step, while each column represents the observations of each feature over all frames. Overall, the tracking system we employ demonstrates that simple, efficient algorithms can track well-localized feature trajectories over multiple frames.  Together with the registration algorithm described in Section \ref{sec:reg}, our complete system could be deployed in low-cost embedded devices.

As a point of comparison, many existing SLAM front-ends employ feature extraction and matching on a frame-by-frame basis \cite{HenryP2010}.  This technique works quite well because RANSAC rejects misaligned features.  However, they are subject to two major drawbacks.  First, real time applications of extract-and-match techniques require hardware acceleration to run in real time.  Second, they match features between frames in feature space, neglecting continuity of spatial observations of these features.

\section{Online 3-D Rigid Body Motion Registration}
\label{sec:reg}

\subsection{Problem Statement}
\label{sec:statement}
First, we shall formulate the 3-D RBMR problem and introduce the notation we will use for the rest of the paper. We denote $\xx_{i,j}\in\Re^3$ as the coordinates of feature $j$ in the $i$th frame, where $i\in[1, \cdots, F]$ and $j\in[1, \cdots, m]$. In the noise-free case, when the same $j$th feature is observed in two different frames $1$ and $i$, its images satisfy a rigid-body constraint:
\begin{equation}
\xx_{i,j} = R_i \xx_{1, j} + T_i \in \Re^3,
\end{equation} 
where $R_i\in\Re^{3\times 3}$ is a rotation matrix and $T_i\in\Re^{3\times 1}$ is a 3-D translation. This relation can be also written in homogeneous coordinates as
\begin{equation}
\xx_{i,j}= \Pi\left[ \begin{matrix} R_i & T_i \\ 0 & 1 \end{matrix}\right] \left[ \begin{matrix}\xx_{1,j}\\ 1 \end{matrix}\right] \doteq \Pi g_i \left[ \begin{matrix}\xx_{1,j}\\ 1 \end{matrix}\right],
\end{equation}
where $\Pi = [I_3, \zero]\in\Re^{3\times 4}$ is a projection matrix.

In the noise-free case, since all the features in the $i$th frame satisfy the same rigid-body motion, one can stack the image coordinates of the same feature in the $F$ frames in a long vector form, and then the collection of all the $m$ features form a data matrix $X$, which can be written as the product of two rank-4 matrices:
\begin{equation}
X\doteq \left[ \begin{smallmatrix}\xx_{1,1} & \cdots & \xx_{1,m}\\
\vdots & \cdots & \vdots \\
\\ \xx_{F,1} & \cdots & \xx_{F,m} \end{smallmatrix}\right]
= \left[ \begin{smallmatrix} \Pi g_1 \\ \vdots \\ \Pi g_F \end{smallmatrix}\right]
\left[ \begin{smallmatrix}\xx_{1,1},& \cdots, & \xx_{1,m} \\ 1, & \cdots, & 1 \end{smallmatrix}\right] \in \Re^{3F\times m}.
\label{eq:rank-4}
\end{equation}
In particular, $g_1=I_4$ represents the identity matrix. It was observed in \cite{TomasiC1992-IJCV,PoelmanC1997-PAMI} that when $F, m \gg 4$, the rank of matrix $X$ that represents a rigid-body motion in space is at most four, which is upper bounded by the rank of its two factor matrices in \eqref{eq:rank-4}. In SfM, the first matrix on the right hand side of \eqref{eq:rank-4} is called a \emph{motion matrix} $M$, while the second matrix is called a \emph{shape matrix} $S$. Although \eqref{eq:rank-4} is not a unique rank-4 factorization of $X$, a canonical representation can be determined by imposing additional constraints on the shape of the object \cite{TomasiC1992-IJCV,PoelmanC1997-PAMI}. 

Lastly, for motion registration, if we denote the 3-D coordinates (e.g., under the world coordinates centered at the camera) of the first frame as: $W_1\doteq [ \xx_{1,1}, \cdots,  \xx_{1,m}] \in \Re^{3\times m}$, then the rigid body motion $(R_i, T_i)$ of the features from the world coordinates to any $i$th frame satisfies the following constraint:
\begin{equation}
W_i\doteq [\xx_{i,1}, \cdots, \xx_{i,m}] = R_i W_1 + T_i\one^T.
\label{eq:registration}
\end{equation}
Using \eqref{eq:registration}, the two transformations $R_i$ and $T_i$ can be recovered by the Orthogonal Procrustes (OP) method \cite{SchonemannP1966}. More specifically, let $\mu_i\in\Re^3$ be the mean vector of $W_i$, and denote $\bar{W}_i$ as the centered feature coordinates after the mean is subtracted. Suppose the SVD of $\bar{W}_i\bar{W}_1^T$ gives rise to:
\begin{equation}
(U, \Sigma, V) = \mbox{svd}(\bar{W}_i\bar{W}_1^T).
\label{eq:procrustes}
\end{equation}
Then the rotation matrix $R_i = UV^T$, and the translation $T_i=\mu_i - R_i\mu_1$.

In this work, we consider an online solution to RBMR. Our goal is to maintain the estimation of a low-rank representation of $X$ and its subsequent new observations $W_i$ with minimal computational complexity. In the rest of the section, we first discuss the initialization step to jump start the low-rank estimation of the initial observations $X$ in Section \ref{sec:initialization}. Then we propose our solution to update the low-rank estimation in the presence of new observations in $i$th frame $W_i$ in Section \ref{sec:online}. Finally, applying our algorithm on real-world data may encounter additional nuisances such as new feature tracks entering the scene and missing data. After the summary of Algorithm \ref{alg}, we will briefly show that the proposed solution can be easily extended to handle these additional conditions in an elegant way.

\subsection{Initialization via Robust PCA}
\label{sec:initialization}
In the initialization step, a robust low-rank representation of $X$ needs to be obtained in the presence of moderate Gaussian noise, data corruption, and outlying image features. The problem can be solved \emph{in closed form} by Robust PCA \cite{CandesE2009-ACM,ZhouZ2010}. Here we model $X\in\Re^{n\times m}$ as the sum of three components:
\begin{equation}
X = L_0 + D_0 + E_0,
\label{eq:decomposition}
\end{equation}
where $L_0$ is a rank-4 matrix that models the ground-truth distribution of the inlying rigid-body motion, $D_0$ is a Gaussian noise matrix that models the dense noise independently distributed on the $X$ entries, and $E_0$ is a sparse error matrix that collects those nonzero coefficients at a sparse support set of corrupted data, outlying image features and bad tracks. 

The matrix decomposition in \eqref{eq:decomposition} can be successfully solved by a \emph{principal component pursuit} (PCP) program:
\begin{equation}
\min_{L, E} \| L \|_* + \lambda \| E \|_1\quad \mbox{subj. to}\quad \|X - L - E\|_F \le \delta,
\label{eq:PCP}
\end{equation}
where $\|\cdot \|_*$ denotes matrix nuclear norm, $\|\cdot \|_1$ denotes entry-wise $\ell_1$-norm for both matrices and vectors, and $\lambda$ is a regularization parameter that can be fixed as $\sqrt{\max(n,m)}$. It has been shown in \cite{CandesE2009-ACM,ZhouZ2010} that when the dimension of matrix $X$ is sufficiently high and with some extra mild conditions on the coefficients of $L_0$ and $E_0$, with overwhelming probability, the global (approximate) solution of $L_0$ and $E_0$ can be recovered. 

The key characteristics of the PCP algorithm are highlighted as follows: Firstly, the regularization parameter $\lambda$ does not necessarily rely on the level of corruption in $E_0$, so long as their occurrences are bounded. Secondly, although the theory assumes the sparse error should be randomly distributed in $X$, the algorithm itself is surprisingly robust to both sparse random corruption and highly correlated outlying features as a small number of column vectors in $X$. Finally, although the original implementation of PCP in \cite{CandesE2009-ACM} is computationally intractable for real-time applications, its most recent implementation based on an augmented Lagrangian method (ALM) has significantly reduced its complexity \cite{LinZ2009}. In this paper, we adopt the ALM solver for Robust PCA, whose average run time is merely a small constant (in general smaller than 20) times the run time of SVD. In our online formulation of SOLO, this calculation only needs to be performed once in the initialization step.

Since the resulting low-rank matrix $L$ may still contain entries of outlying features, an extra step needs to be taken to remove those outliers. In particular, one can calculate the $\ell_0$-norm of each column in $E_0 = [\ee_1, \ee_2, \cdots, \ee_m]$. With respect to an outlier threshold $\tau$, if $\|\ee_i\|_0>\tau$, then $\ee_i$ represents dense corruption on the corresponding feature track and hence should be regarded as an outlier.\footnote{For those coefficients in $\ee_i$ with small nonzero values, a hard-thresholding can be applied to reduce the values to zero.} Subsequently, the indices of the inliers define a support set $I\subset [1, \cdots, m]$.
Hence, we denote the cleaned low-rank data matrix after outlier rejection as 
\begin{equation}
\hat{L}\doteq L^{(I)}.
\label{eq:outlier-rejection}
\end{equation}

Finally, we note that although in \eqref{eq:PCP}, $L$ represents the optimal matrix solution with the lowest possible rank, due to additive noise and data corruption in the measurements, its rank may not necessarily be less than five. Therefore, to enforce the rank constraint in the RBMR problem and further obtain a representative of the shape matrices that span the 4-D subspace, an SVD is performed on $\hat{L}$ to identify its right eigenspace:
\begin{equation}
(U, \Sigma, V) = \mbox{svds}(\hat{L}, 4),
\label{eq:svds}
\end{equation}
where $V^T\in\Re^{4\times m}$ is then a representative of the rigid body's shape matrices.

\subsection{Sparse Online Low-rank projection and Outlier rejection (\acronym)}
\label{sec:online}

In this section, we propose a novel algorithm that projects new observations $W_i$ from the $i$th frame onto the rigid-body shape subspace. This subspace is parameterized by the shape matrix $V^T$ that we have estimated in the initialization step.\footnote{In this paper, we may choose to abuse the notation of $V^T$ to also represent the 4-D subspace.} Traditionally, a (least squares) subspace projection operator would project a (noisy) sample perpendicular to the surface of the subspace that it is close to, which only involves basic matrix-vector multiplication. However, in anticipation of continual random feature corruption during the course of feature tracking for RBMR, the projection must also be robust to sparse error corruption in $W_i$. Hence, we contend that {\acronym} is a more appropriate yet still efficient algorithm to achieve online motion registration update.

Given the initialization $\hat{L}$ and the inlier support set $I$, without loss of generality, we assume $W_i$ only contains those features in the support set $I$. As discussed in \eqref{eq:rank-4} and \eqref{eq:svds}, matrix $V^T$ from the SVD of $\hat{L}$ is a representative of the class of all the shape matrices of the rigid body up to an ambiguity of 4-D rotation on the subspace. Therefore, the new observations $W_i$ of the same features should also lie on the same shape subspace. That is, let $W_i = [\ww_1^T; \ww_2^T; \ww_3^T]$, where each $\ww_1^T\in\Re^{1\times m}$ is a row vector. Then
\begin{equation}
\ww_j^T = \aa^T V^T\quad \mbox{for some}\quad \aa^T\in\Re^{1\times 4}.
\end{equation}

\begin{figure}[tbp!]
  \centering
  \includegraphics[width=0.4\textwidth]{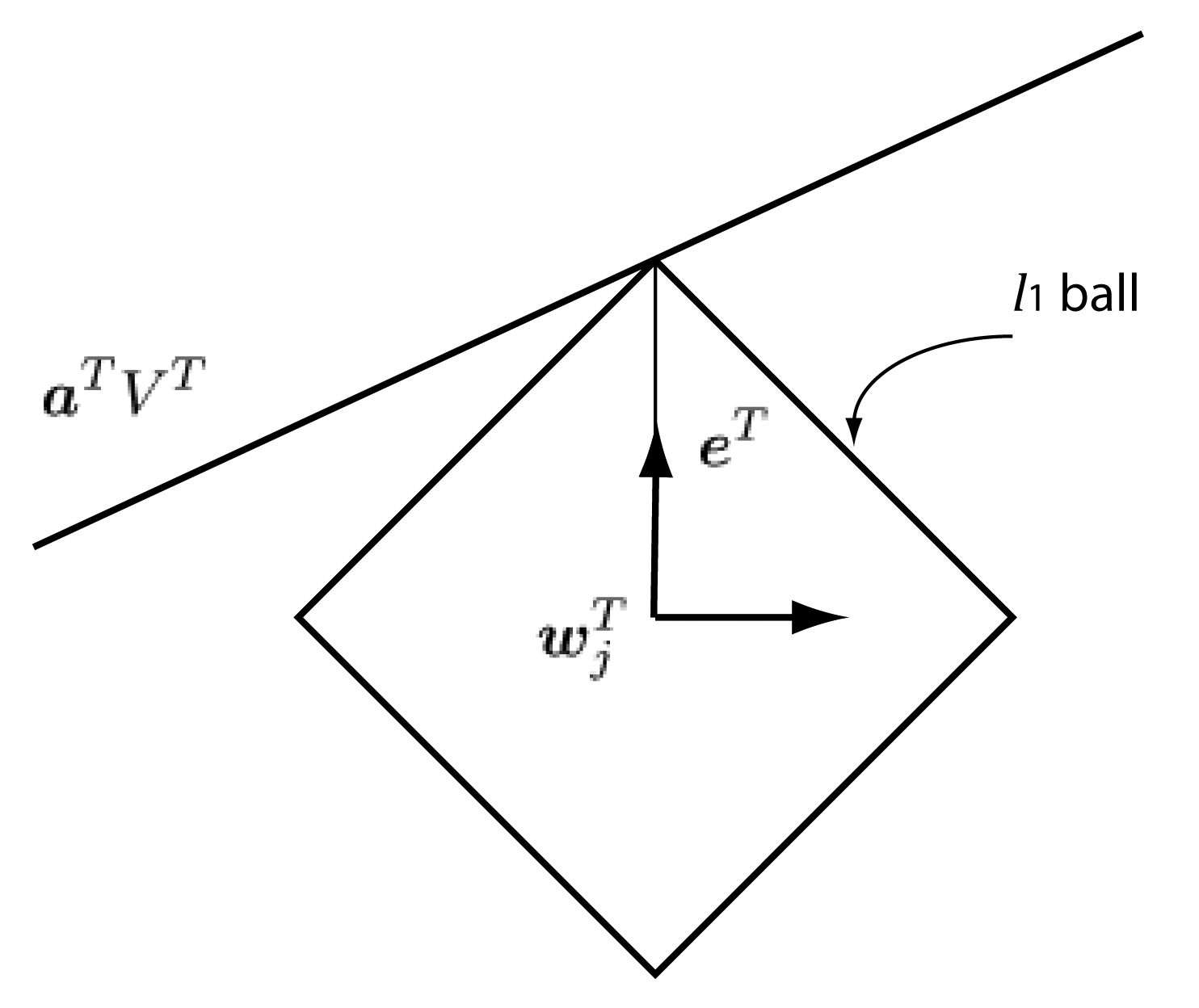}
  \caption{\small A visualization of sparse subspace projection as basis-pursuit denoising, which can be solved by $\ell_1$-minimization.}
  \label{fig:BPDN}
\end{figure}

In the presence of sparse corruption, the row vector $\ww_j^T$ is perturbed by a sparse vector $\ee$:
\begin{equation}
\ww_j^T = \aa^T V^T + \ee^T, \quad \mbox{where } \ee^T \in\Re^{1\times m}.
\label{eq:row-sparse-projection}
\end{equation}
The sparse projection constraint \eqref{eq:row-sparse-projection} bears resemblance to \emph{basis-pursuit denoising} (BPDN) in compressive sensing literature \cite{ChenS1998}, as a sparse error perturbs a high-dimensional sample away from a low-dimensional subspace model. The standard procedure of BPDN using $\ell_1$-minimization ($\ell_1$-min) is illustrated in Figure \ref{fig:BPDN}.

\begin{figure}[tbp!]
  \centering
  \includegraphics[width=0.4\textwidth]{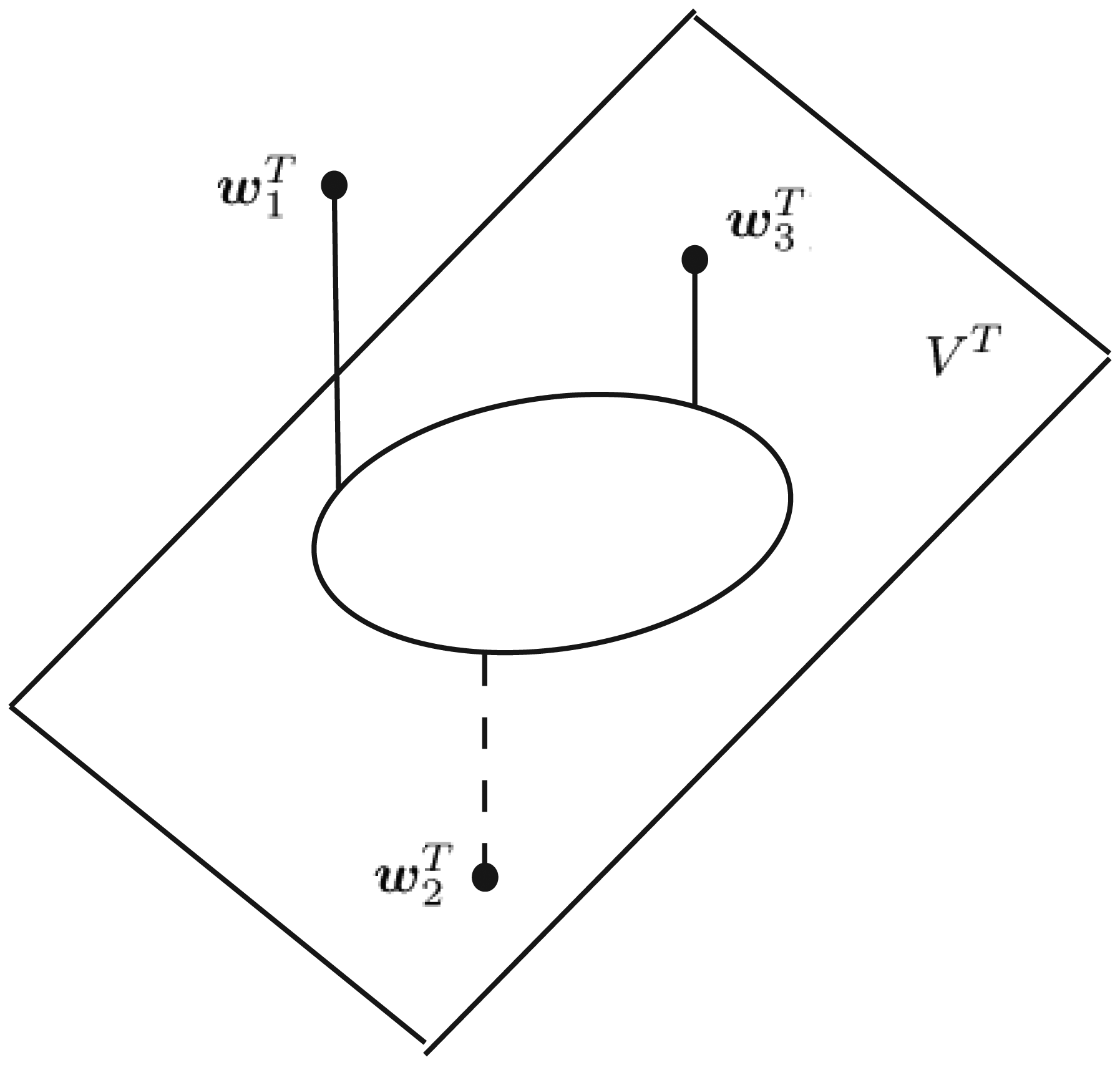}
  \caption{\small The row vectors of $W$ should be projected onto a manifold in $V^T$ that represents a valid rigid-body motion.}
  \label{fig:subspace-projection}
\end{figure}

However, we notice that a BPDN-type solution via $\ell_1$-min may not be the optimal solution to our problem. The reason is that the row vectors in $W = [\ww_1^T; \ww_2^T; \ww_3^T]$ are not three arbitrary vectors in the 4-D subspace $V^T$. In fact, the three vectors must be projected onto a nonlinear manifold $M$ embedded in the shape subspace $V^T$, and the span of the shape model can be interpreted as the linear hull of the feasible rigid-motion motions between $W_1$ and $W_i$. Figure \ref{fig:subspace-projection} illustrates this rigid-body constraint applied to sparse subspace projection in 3-D.

Our algorithm of \emph{sparse shape subspace projection} is described as follows. Given the observation $W_i$ and a shape subspace $V^T$, the algorithm minimizes:
\begin{equation}
\min_{E, A} \| E\|_1\quad \mbox{subj. to}\quad W_i = AV^T + E.
\label{eq:subspace-projection}
\end{equation}
By virtue of low dimensionality of this hull, together with the sparsity of the residual, the projected data $AV^T$ should be well localized on the manifold. Hence, in addition to being consistent with a realistic (sparse) noise model, the new sparse subspace projection algorithm \eqref{eq:subspace-projection} also implies the benefit of good localization in the motion space.

The objective can be solved quite efficiently (and much faster than solving RPCA in the initialization) by the same augmented Lagrangian approach in \cite{LinZ2009}:
\begin{equation}
\min_{A, E, Y, \mu} \| E \|_1 + \langle Y, W_i-AV^T-E \rangle + \frac{\mu}{2} \| W_i - AV^T - E\| _F^2,
\end{equation}
where $Y$ is a matrix of Lagrange multipliers, and $\mu>0$ represents a monotonically increasing penalty parameter during the optimization. The optimization only involves a soft-thresholding function applied to the entries of $E$ and matrix-matrix multiplication for the update of $A$ and $E$, and does not involve computation of singular values as in RPCA.

Finally, the rigid-body motion between each $W_i$ and the first reference frame $W_1$ after the projection can be recovered by the OP algorithm \eqref{eq:procrustes}. However, as the projection \eqref{eq:subspace-projection} may be also affected by dense Gaussian noise, the estimated low-rank component may not accurately represent a consistent rigid-body motion. As a result, what we can do is to identify an index set $I_i$ for those uncorrupted features with zero coefficients in $E$. The OP algorithm will be applied only using the uncorrupted \emph{original} features in $W_1$ and $W_i$. In a sense, this motion registration algorithm resembles the strategy in RANSAC to select inlying sample sets. However, our algorithm has the ability to directly identify the corrupted features via sparse subspace projection, and hence the process is noniterative and more efficient.

The complete algorithm, \emph{Sparse Online Low-rank projection and Outlier rejection} (SOLO), is summarized in Algorithm \ref{alg}.
\begin{algorithm}[ht!]
\caption{\acronym}
\label{alg}
{\bf Input:} Initial observations $X$, feature coordinates of the reference frame $W_1$, and $W_i$ for each subsequent frame $i$.
\begin{algorithmic}[1]
\STATE {\bf Init:} Compute $L$ and $I$ of $X$ via RPCA \eqref{eq:PCP}.
\STATE $W_1\leftarrow W_1^{(I)}$, remove outliers in the reference frame.
\STATE $[U, \Sigma, V] = \mbox{svds}(L^{(I)}, 4)$.
\FOR{Each new observation frame $i$}
\STATE  $W_i\leftarrow W_i^{(I)}$.
\STATE Identify corruption $E$ via sparse subspace projection \eqref{eq:subspace-projection}.
\STATE Let $I_i$ be the index set of uncorrupted features in $W_i$.
\STATE Estimate $(R_i, T_i)$ using inlying samples in $I_1\cap I_i$.
\ENDFOR
\end{algorithmic}
{\bf Output:} Inlier support set $I$, rigid-body motions $(R_i, T_i)$.
\end{algorithm}

Before we proceed to discuss results from our experiment, it is worth mentioning a straightforward yet elegant extension of the algorithm in the presence of missing data. In the initialization step, one can rely on a variant of RPCA to recover the missing data in matrix $X$. The technique is known as \emph{low-rank matrix completion} \cite{CaiJ2008,CandesE2009-ACM}, which minimizes a similar low-rank representation objective constrained on the observable coefficients:
\begin{equation}
\min_{L, E} \|L\|_* + \lambda \|E \|_1 \quad \mbox{subj. to} \quad \mathcal{P}_\Omega(L + E) = \mathcal{P}_\Omega(X),
\label{eq:matrix-completion}
\end{equation}
where $\Omega$ is an index set of those features that remain visible in $X$, and $\mathcal{P}_\Omega$ is the orthogonal projection onto the linear space of matrices supported on $\Omega$.

Using low-rank matrix completion \eqref{eq:matrix-completion}, in the presence of a partial measurement of new feature tracks, those incomplete new observations should be identified as tracks with missing data. Then a new initialization step using \eqref{eq:matrix-completion} should be performed on a new data matrix $X$ that includes the new tracks to re-establish the shape subspace and inlier support set $I$ as in \eqref{eq:svds}.

\section{Experiment}
\label{sec:exp}
In this section, we validate the performance of SOLO algorithm and compare with the classical RANSAC solutions, which has been the most popular solution to date for SLAM and motion registration. In the rest of the section, the two algorithms will be applied to a thorough list of simulations and a real-world experiment. The benchmarks are calculated on a 2\,GHz PC with an Intel Core i7 processor and in MATLAB environment.

\subsection{Simulated Analysis}
We first use synthesized data to benchmark the accuracy and speed of our batch motion registration algorithm described in Section \ref{sec:initialization}.
The calculation of $(R_i, T_i)$ between each pair of $W_1$ and $W_i$ will be based on $\hat{L}$ alone as the output of RPCA and outlier rejection \eqref{eq:outlier-rejection}. We compare the performance of motion registration by RPCA with that by the classical solution of RANSAC on a frame-by-frame basis. The minimal feature set in RANSAC is set to four. 

\begin{figure}[th!]
  \centering
  \subfloat[\scriptsize Added data corruption $D_0 + E_0$]{\includegraphics[width=0.3\textwidth]{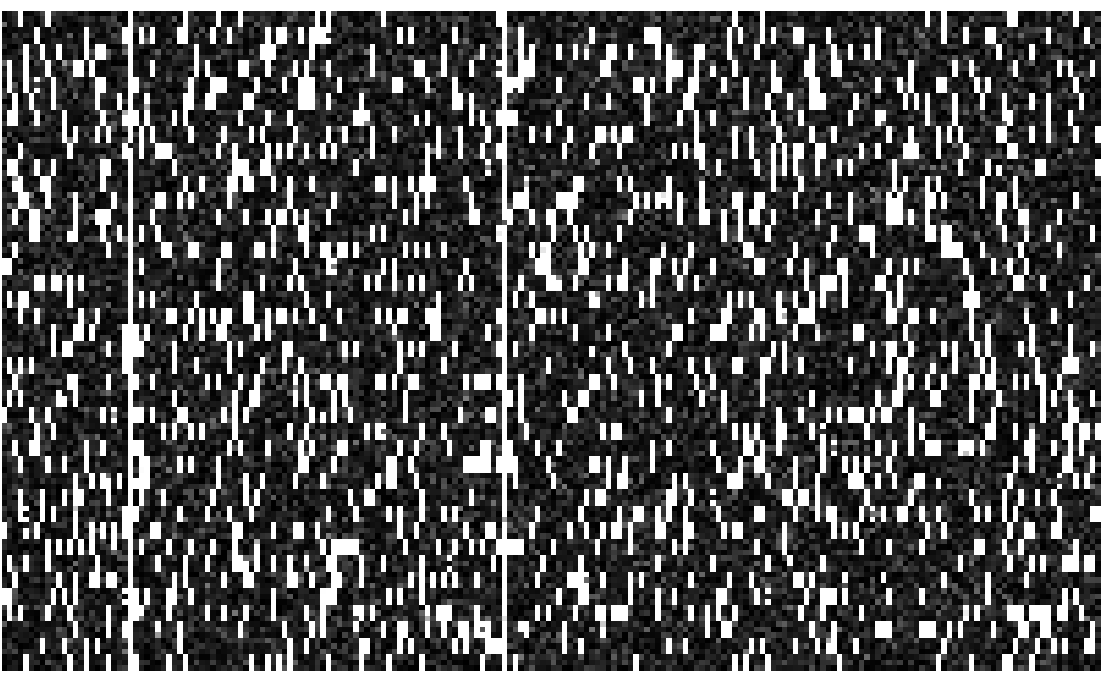}} 
  \hspace{2pt}
  \subfloat[\scriptsize Rejected features by RANSAC]{\includegraphics[width=0.3\textwidth]{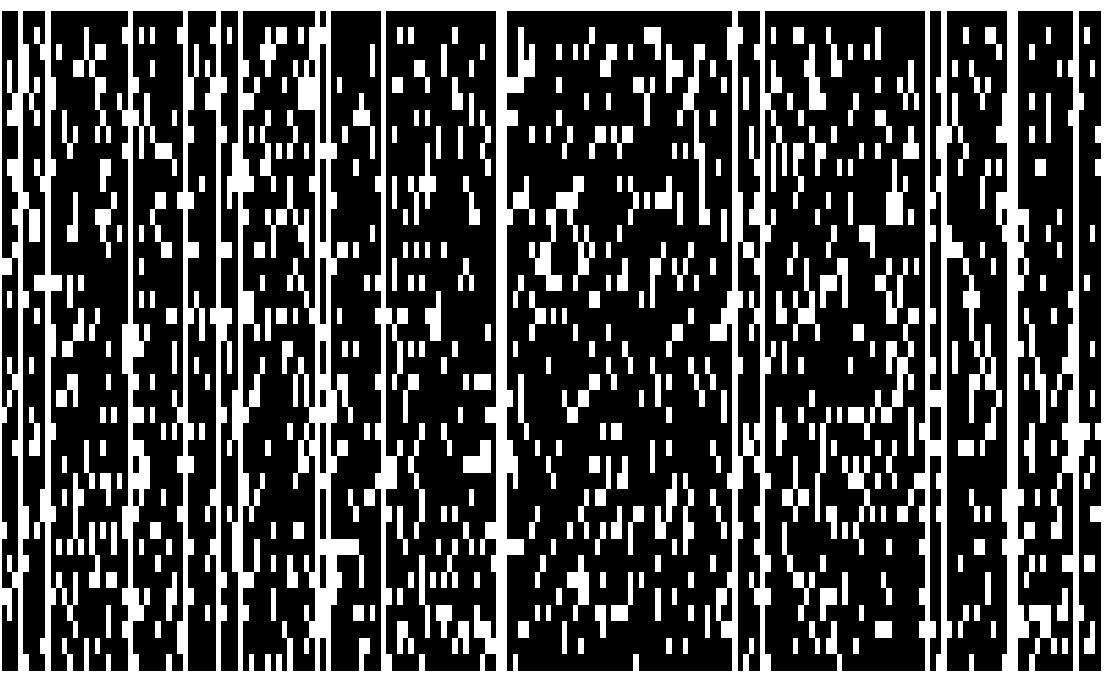}}\\
  \hspace{2pt}
    \subfloat[\scriptsize Estimated sparse error by RPCA]{\includegraphics[width=0.3\textwidth]{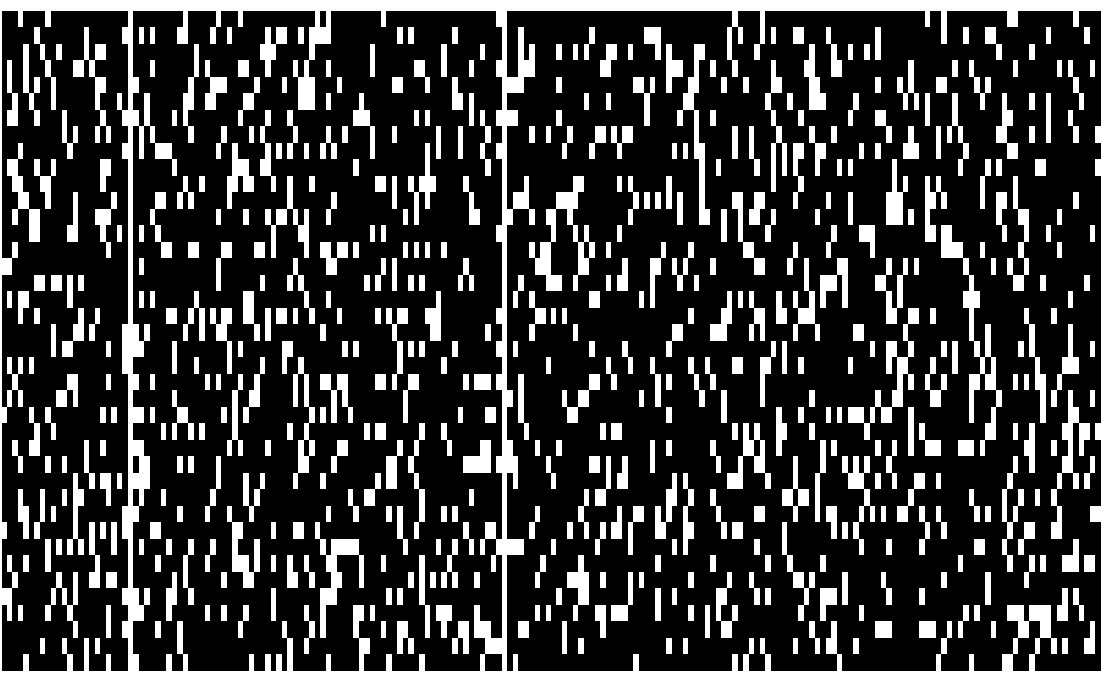}}
  \hspace{2pt}
  \subfloat[\scriptsize Sparse error difference $\abs{E_0 - E^*}$]{\includegraphics[width=0.3\textwidth]{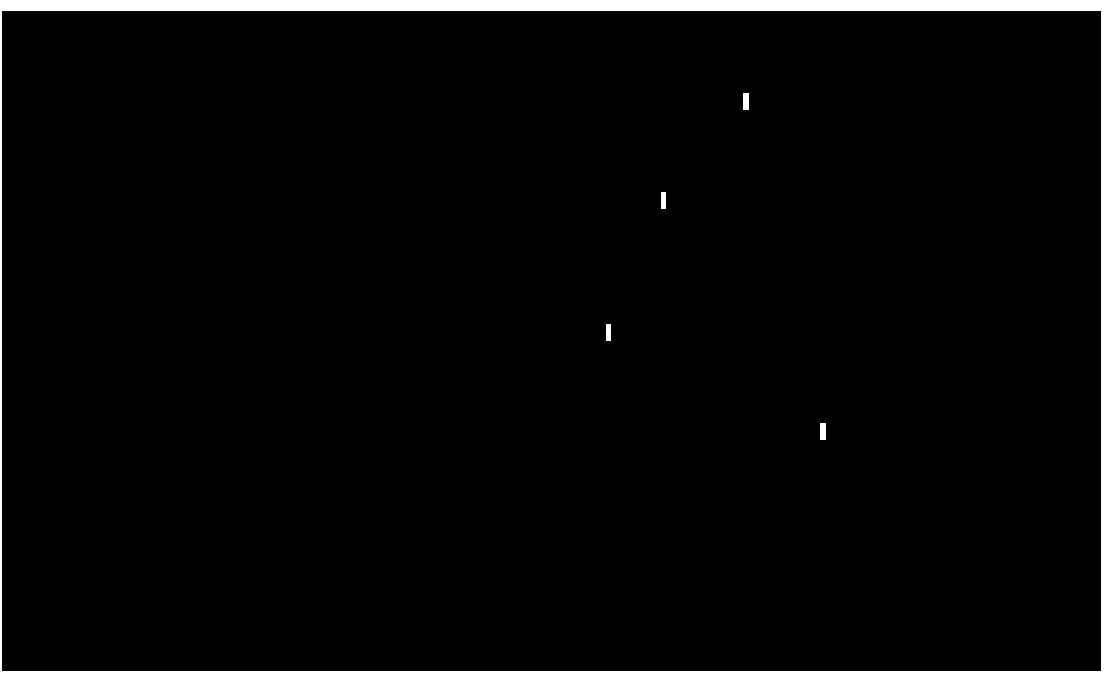}}
  \hspace{2pt}
  \subfloat[\scriptsize Ground-truth difference $\abs{L_0 - L^*}$\label{fig:simulation-e}]{\label{fig:tiger}\includegraphics[width=0.3\textwidth]{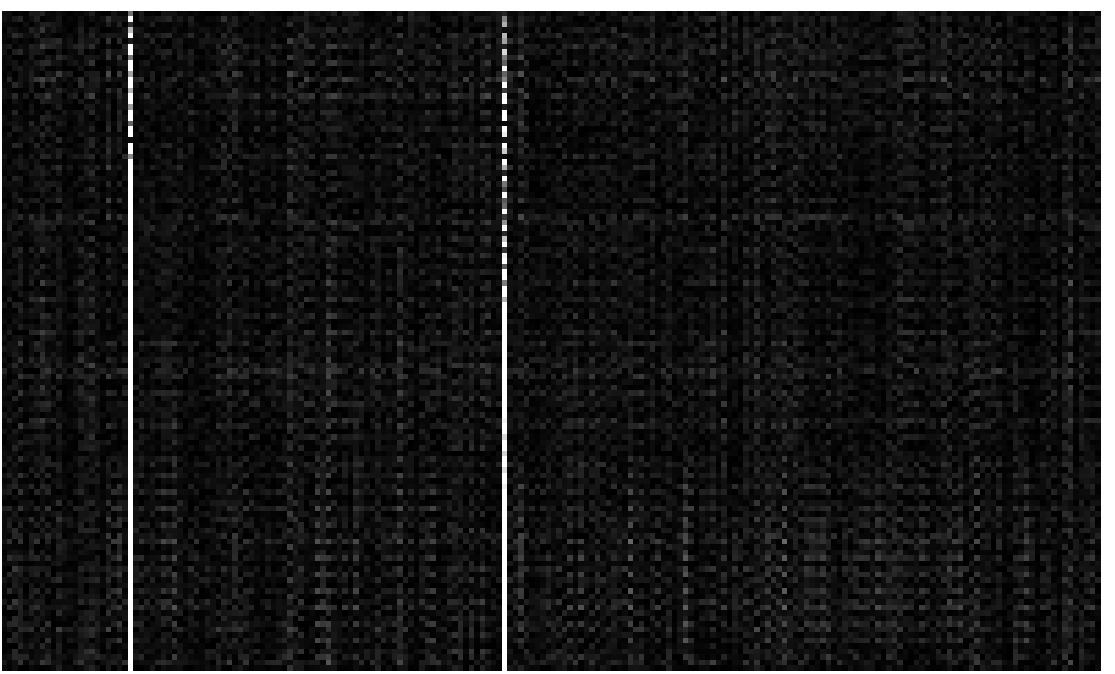}}
  \caption{\small A visualization of the estimation error of a simulated motion matrix $X$ by RANSAC and RPCA. The added data corruption and estimation difference are represented as white pixels in the images. }
  \label{fig:simulation}
\end{figure}

In one simulation, the outlier rejection results in motion registration by RPCA and RANSAC are visualized in Figure \ref{fig:simulation}. In this example, we observe that RPCA is much more effective in identifying both random data corruption and outlying feature tracks (that post inconsistent feature measurements in the entire columns) than RANSAC. Also note that the large coefficient difference in the two columns of Figure \ref{fig:simulation-e} should not be a concern, as it is well known that RPCA cannot uniquely recover dense column corruption \cite{CandesE2009-ACM}, and nevertheless the corresponding features will be rejected as outliers by \eqref{eq:outlier-rejection}. Finally, we can also see quite significant difference between the ground-truth low-rank matrix $L_0$ and its estimate $L^*$. It shows the accuracy of RPCA is still sensitive to high variance dense Gaussian noise.

\begin{figure}[th!]
  \centering
\subfloat[\scriptsize Average runtime vs. corruption percentage]{\label{fig:runtime}\includegraphics[width=0.4\textwidth]{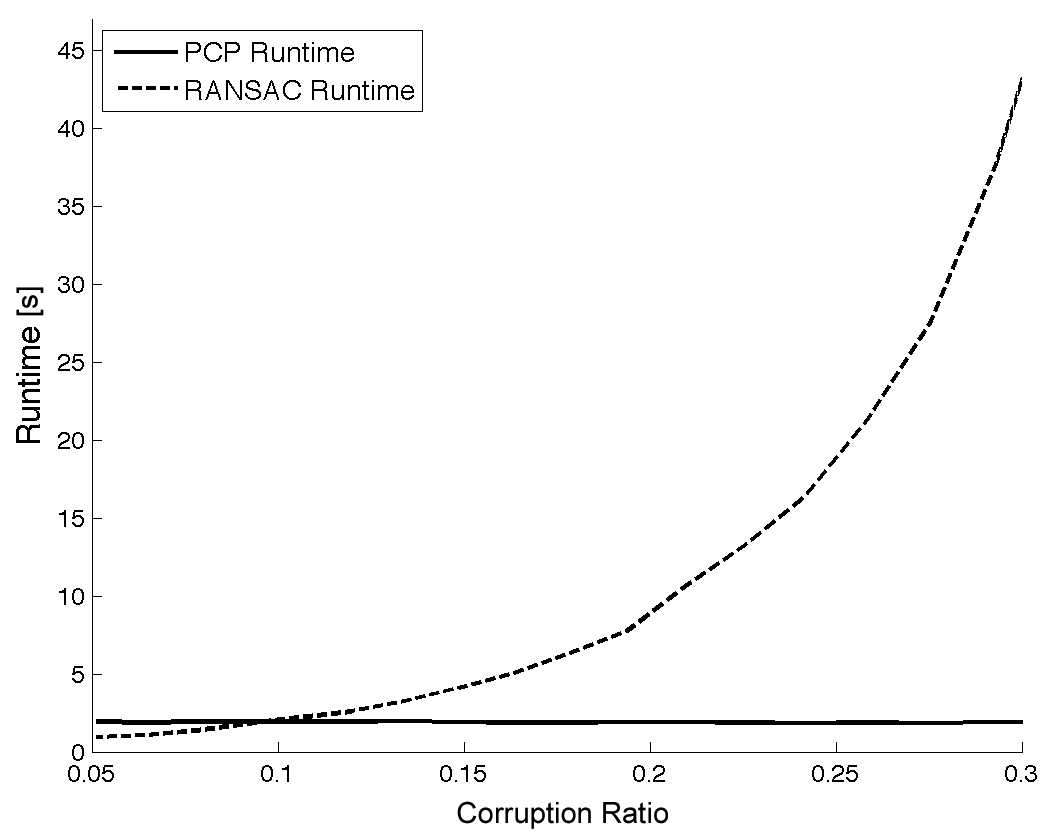}}
\subfloat[\scriptsize RPCA runtime]{\label{fig:compfeats}\includegraphics[width=0.4\textwidth]{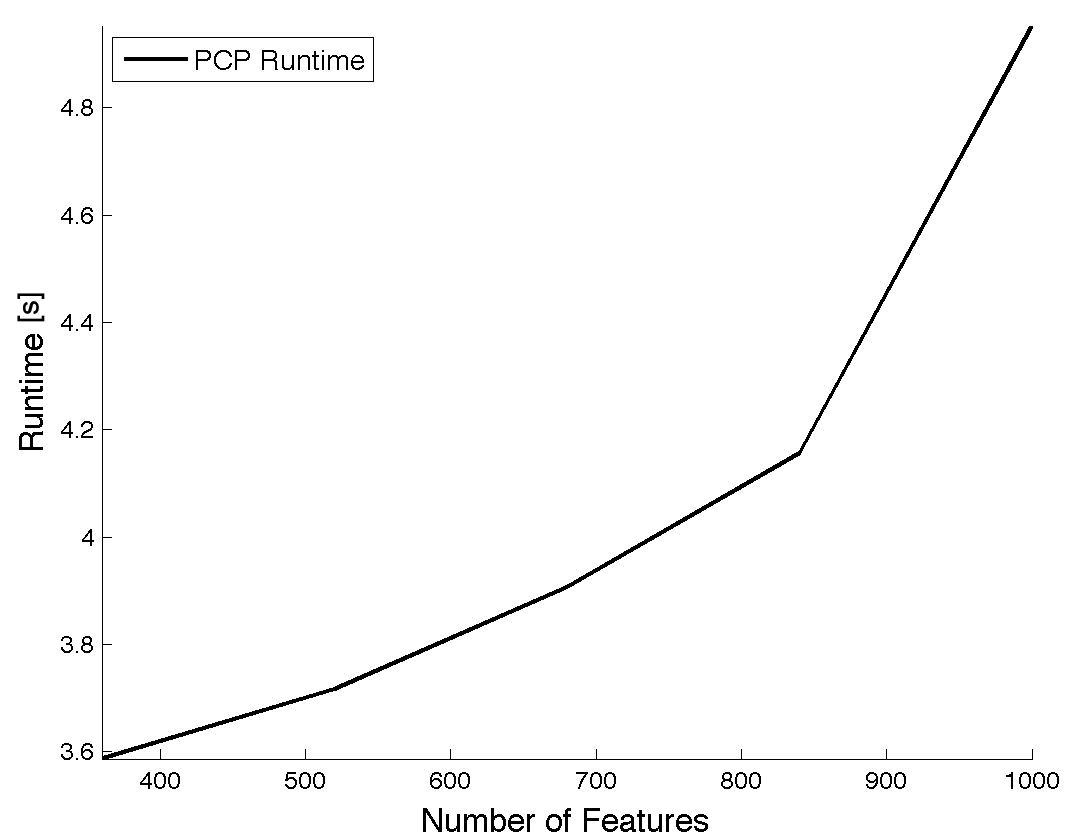}}\\
\subfloat[\scriptsize RPCA runtime vs. number of frames]{\label{fig:compframes}\includegraphics[width=0.4\textwidth]{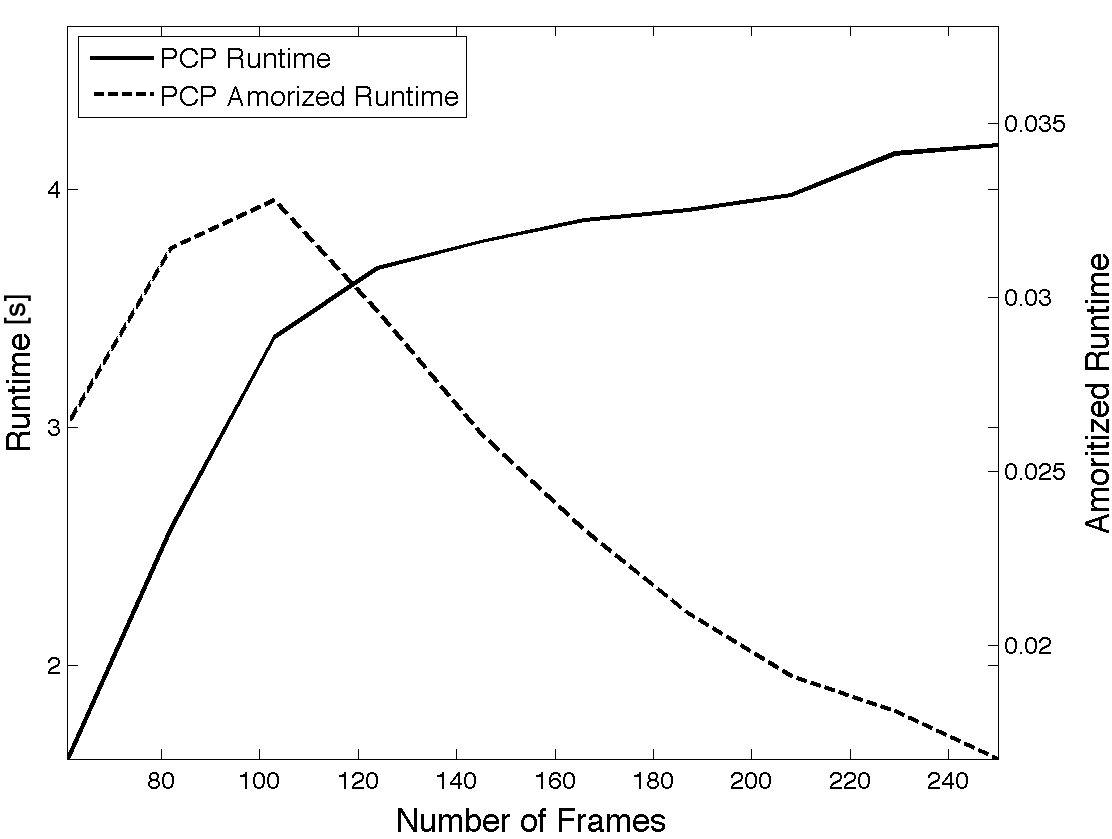}}
\subfloat[\scriptsize Motion registration accuracy vs. noise variance]{\label{fig:noise-accuracy}\includegraphics[width=0.4\textwidth]{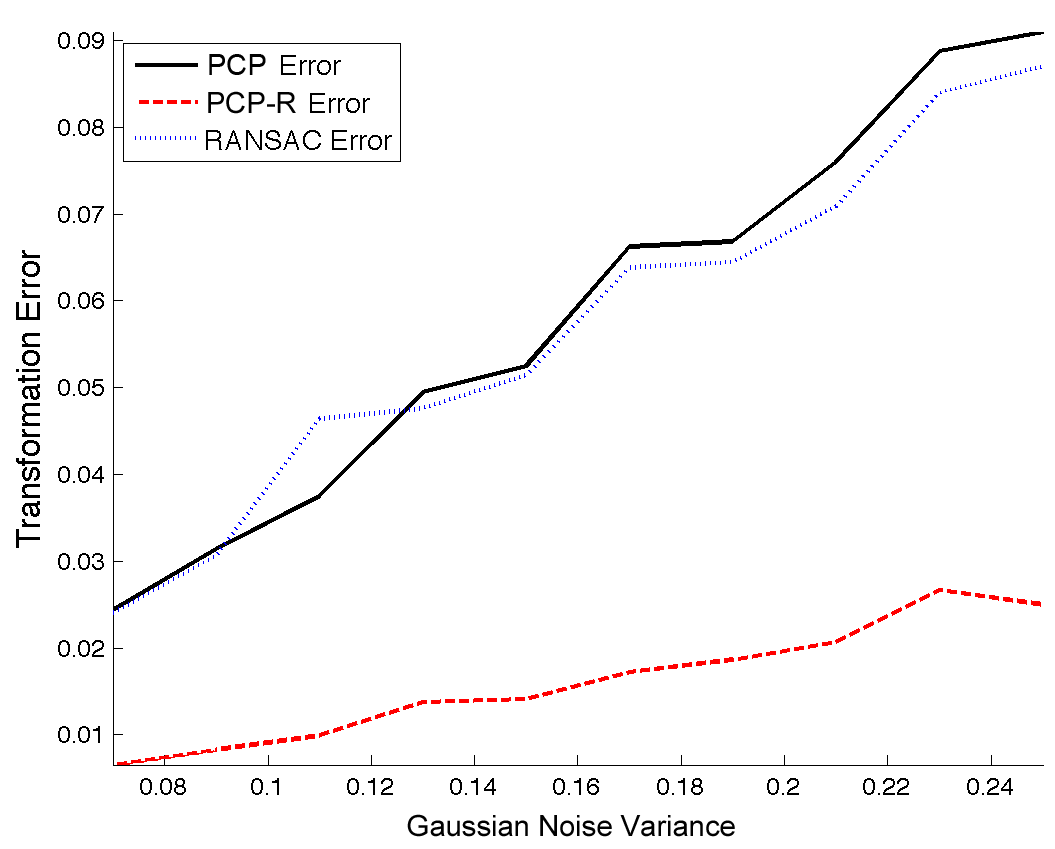}}
\caption{\small A simulated comparison between RPCA and RANSAC. PCP is based on the ALM method. PCP-R means the RPCA approach with a RANSAC-style iterative refinement stage.}
\label{fig:benchmark}
\end{figure}

To overcome the issue of dense Gaussian noise in RPCA, our recommended implementation further adds a RANSAC-style refinement stage, which selects a minimal set of inlying samples from the support set already identified by RPCA. Correspondences consistent with the constructed model are merged until the refinement stage converges. Typically this recursive refinement process converges in 2--4 iterations. With this in mind, we show the accuracy and speed of motion registration using RPCA and RANSAC in Figure \ref{fig:benchmark}. In Figure \ref{fig:noise-accuracy} the motion registration accuracy with respect to two matrices $(R, T)$ is measured by the sum of the difference to the ground truth $(R_0, T_0)$ in Frobenius norm.

We can see in Figure \ref{fig:runtime}, with certain level of accuracy confidence, the average runtime of RANSAC grows superlinearly with the increase of the corruption percentage, while RPCA remains effective in compensating those corruptions in the low-rank matrix. Figure \ref{fig:compfeats} and \ref{fig:compframes} show reasonable increase in computation time for RPCA with respect to the number of features and the number of frames in the motion window $X$. Finally, the accuracy about the estimated rigid-body transformation is shown in Figure \ref{fig:noise-accuracy}. Without the additional refinement stage, RPCA already achieves comparable result than RANSAC. If the iterative refinement is added to the algorithm, we can see significant improvement in the estimation of the motion. Notice that the estimation errors of $R$ and $T$ are already very small in all three cases, as shown on the $y$-axis.

\subsection{Performance on Kinect Data}

\begin{figure}[p]
  \centering
  \subfloat[\scriptsize RANSAC reconstruction]{\includegraphics[width=0.23\textwidth,height=0.15\textheight]{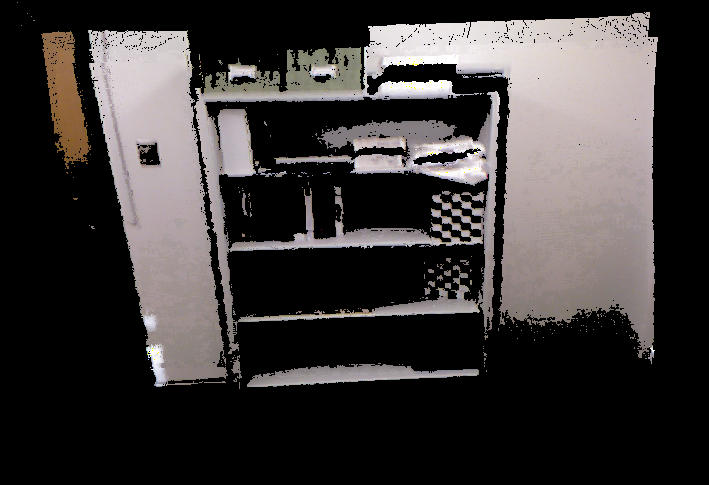}} 
  \hspace{2pt}
  \subfloat[\scriptsize SOLO reconstruction]{\includegraphics[width=0.23\textwidth,height=0.15\textheight]{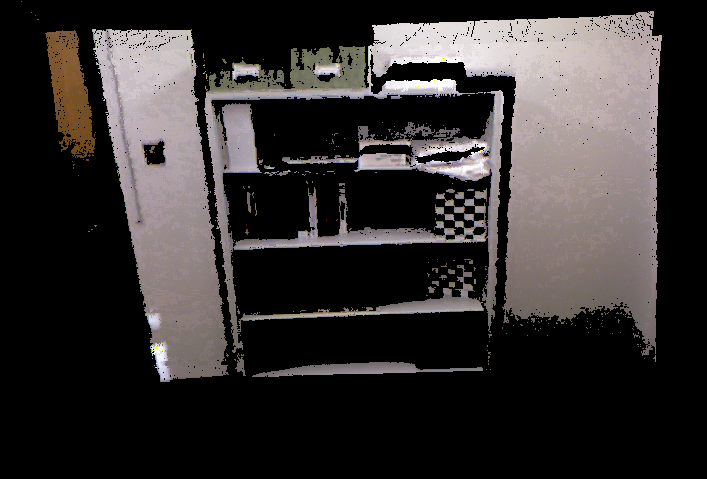}} \\
  \subfloat[\scriptsize RANSAC checkerboard detail]{\includegraphics[width=0.23\textwidth,height=0.15\textheight]{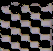}} 
  \hspace{2pt}
  \subfloat[\scriptsize SOLO checkerboard detail]{\includegraphics[width=0.23\textwidth,height=0.15\textheight]{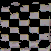}} \\
  \subfloat[\scriptsize RANSAC feature registration]{\includegraphics[width=0.45\textwidth,height=0.15\textheight]{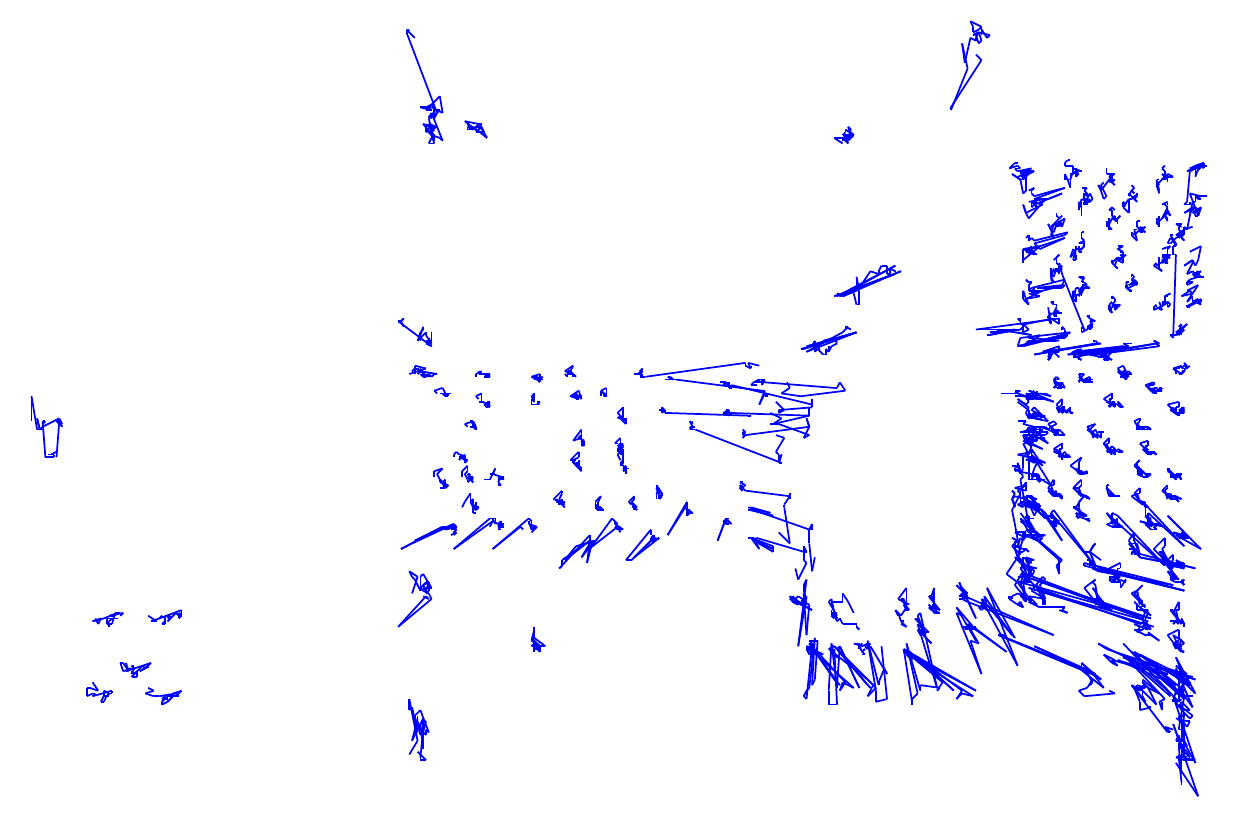}} \\
  \subfloat[\scriptsize SOLO feature registration]{\includegraphics[width=0.45\textwidth,height=0.15\textheight]{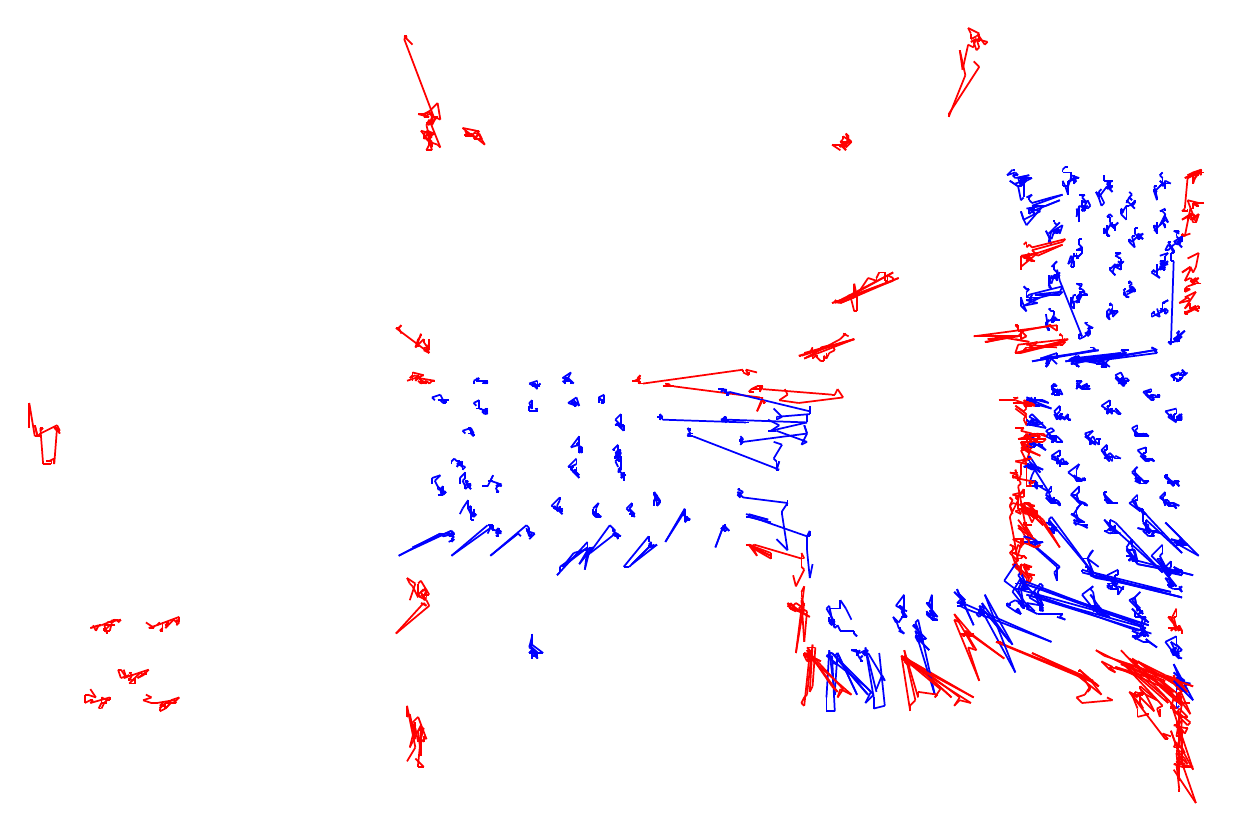}} \\
  \subfloat[\scriptsize SOLO feature registration with refinement]{\includegraphics[width=0.45\textwidth,height=0.15\textheight]{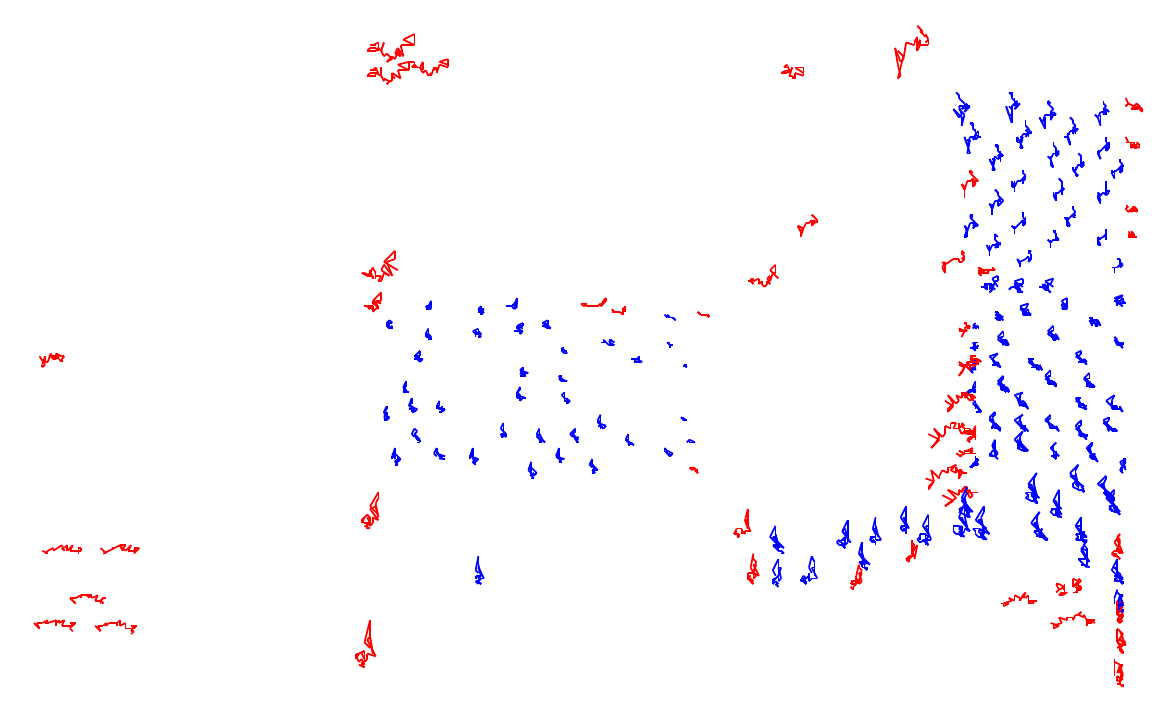}} \\
  \caption{A comparison of SOLO- and RANSAC-based registration results for a real world 3-D reconstruction problem. (e)-(g) are feature trajectories in the world coordinate system. Features discarded by our algorithm are shown in red.}
  \label{fig:experiment}
\end{figure}

We now test the performance of the online SOLO algorithm combined with the KLT tracker on a set of real-world depth data collected by a Microsoft Kinect sensor. The data are collected in an indoor lab environment. The motion registration and scene reconstruction results are shown in Figure \ref{fig:experiment}.

In our experiment, we found the KLT tracking scheme applied on Kinect to be highly effective in practice, producing upwards of two hundred tracks in a typical indoor setting.  This ensures that the initialization of $X$ has enough features to converge to the correct low-rank model $L^{*}$. As expected, the KLT feature tracker also produces small amounts of local jumps due to repetitive object textures (e.g., the checkerboard pattern).

For this experiment, we have tuned RANSAC specifically for the empirical sample corruption ratio in the scene. Despite this effort, SOLO is still faster by a factor of two compared to RANSAC. We emphasize that oracle tuning provides a lower bound on the complexity of RANSAC, and its complexity would be much higher in a less-controlled, online setting.  

The enlarged checkerboard references demonstrate crisper results for the SOLO registration than the RANSAC registration. More interestingly, Figs. 6e-g demonstrate feature registrations for RANSAC and SOLO.  The red trajectories are those which are selected by SOLO for rejection. Despite spurious recorded behavior, such as coarse spatial discontinuities, many of the tracks are salvageable and properly localized in the cleaned data $L^{*}$. Overall, SOLO demonstrates equally good or better registration quality than RANSAC, if measured qualitatively.

\section{Conclusion and Discussion}\label{sec:con}

We have proposed an online 3-D motion registration algorithm called SOLO. Its main advantage compared to existing robust statistical methods such as RANSAC is that the algorithm is capable of exploiting the underlying low-rank matrix structure in describing the motion and shape of a dominant rigid-body. The initialization step employs Robust PCA to recover such low-rank matrices and compensate gross feature corruption and outliers. The online update step sequentially projects new observations onto the inlier shape subspace by a sparse subspace projection technique, which is efficient to implement as a convex program. In our extensive experiment, we have demonstrated equally good or better motion registration accuracy compared to RANSAC, with significant speed-up by one to two orders of magnitude.

For future problems, the convincing results shown in the paper can bring SOLO to a broader range of applications in SLAM. In this paper, we have considered the motion registration problem for a single motion. In a more complex dynamic scene, multiple motions may be captured by the 3-D camera. In addition, the multiple motions may be either independent or constrained (e.g., a humanoid robot consists of multiple linked rigid limbs and the torso). These are some of the interesting problems we intend to investigate further. We believe the SOLO framework has laid a solid foundation for us to tackle these problems.

\end{document}